\RequirePackage{amsthm}
\documentclass[sn-mathphys,Numbered]{sn-jnl}

\usepackage{graphicx}%
\usepackage{multirow}%
\usepackage{amsmath,amssymb,amsfonts}%
\usepackage{amsthm}%
\usepackage{mathrsfs}%
\usepackage[title]{appendix}%
\usepackage{xcolor}%
\usepackage{subfigure}%
\usepackage{textcomp}%
\usepackage{manyfoot}%
\usepackage{tikz}
\usetikzlibrary{arrows}
\usetikzlibrary{patterns}
\usepackage{booktabs}%
\usepackage{algorithm}%
\usepackage{algorithmicx}%
\usepackage{algpseudocode}%
\usepackage{listings}%

\theoremstyle{thmstyleone}%
\theoremstyle{thmstyletwo}%
\newtheorem{example}{Example}%
\newtheorem{remark}{Remark}%

\theoremstyle{thmstylethree}%
\newtheorem{definition}{Definition}%

\begin{document}

\title[Epistemic uncertainty in data-driven strategies]{Representations of epistemic uncertainty and awareness in data-driven strategies}


\author*[1,2]{\fnm{Mario} \sur{Angelelli}}\email{mario.angelelli@unisalento.it}

\author[1]{\fnm{Massimiliano} \sur{Gervasi}}\email{massimiliano.gervasi@unisalento.it}

\affil*[1]{\orgname{University of Salento},
\orgdiv{CAMPI - Centre of Applied Mathematics and Physics for Industry}
\orgaddress{
\city{Lecce}, \postcode{73100}, 
\country{Italy}}
}

\affil[2]{\orgname{INdAM - Istituto Nazionale di Alta Matematica}, \orgdiv{GNSAGA}, 
\orgaddress{
\country{Italy}}
}


\abstract{
The diffusion of AI and big data is reshaping decision-making processes by increasing the amount of information that supports decisions while reducing direct interaction with data and empirical evidence. This paradigm shift introduces new sources of uncertainty, as limited data observability results in ambiguity and a lack of interpretability.
The need for the proper analysis of data-driven strategies motivates the search for new models that can describe this type of bounded access to knowledge. 

This contribution presents a novel theoretical model for uncertainty in knowledge representation and its transfer mediated by agents. We provide a dynamical description of knowledge states by endowing our model with a structure to compare and combine them. Specifically, an update is represented through combinations, and its explainability is based on its consistency in different dimensional representations. 

We look at inequivalent knowledge representations in terms of multiplicity of inferences, preference relations, and information measures. Furthermore, we define a formal analogy with two scenarios that illustrate non-classical uncertainty in terms of ambiguity (Ellsberg's model) and reasoning about knowledge mediated by other agents observing data (Wigner's friend). Finally, we discuss some implications of the proposed model for data-driven strategies, with special attention to reasoning under uncertainty about business value dimensions and the design of measurement tools for their assessment.
}




\maketitle

\section{Introduction}
\label{sec: introduction} 

The ongoing technological evolution enables the generation, acquisition, storage, and analysis of an ever-increasing amount of data. In this context, data can be considered raw resources that need to be manipulated, transformed, and combined to extract usable information and knowledge \citep{Ackoff1989,YlijokiPorras2019}. In the literature, data, especially big data, are described in terms of \textit{Velocity}, \textit{Variety}, and \textit{Volume} \citep{Laney} using the original \textit{V's of the big data} (hereafter referred to as \textit{V's}). Since 2010, this definition of big data associated with a multi-dimensional characterisation has been enriched by new V's \citep{Hussien}. However, some of these features are not intrinsic; namely, they also depend on factors external to the data. Furthermore, their evaluation can be affected by a certain degree of subjectivity. The lack of a context that specifies such characteristics according to the data-driven strategy has occasionally led to confusing definitions \citep{YlijokiPorras2016}, making it challenging to identify which V's to accept \citep{PatgiriAhmed2016}. In particular, \textit{Value} is listed among the V's but is also considered a big data feature distinguished from the data features \citep{Uddin_Gupta2014, PatgiriAhmed2016}, as it can be defined according to the other V's \citep{Geerts_O'Leary2002} and is more oriented to the way the analysed data will be used \citep{YlijokiPorras2019}. According to these definitions, we refer to \emph{data-driven strategies} as a process of data analysis supported by Artificial Intelligence (AI) to create \textit{strategic value} in terms of reusable knowledge and decision support \cite{Vitari_Raguseo_2020}.

\subsection{Motivations}
\label{subsec: motivation}

In this context, the focus on big data analysis and the strategic value generated by data-driven strategies should be investigated according to the Knowledge View, as formalised in the Data-Information-Knowledge-Wisdom (DIKW) paradigm \citep{Ackoff1989}. This model has been extended to the Raw Data-Data Formats-Information-Knowledge-Wisdom (RDIKW) paradigm proposed by \citet{Wu2022707} and to the Big Data Value Graph, where the RDIKW model is integrated with the Data Mesh \citep{Gervasi2023}.
To support the identification of value generation through data-driven strategies, specific methods have been defined, both in terms of conceptual frameworks and measurement tools. The former encompass value-dimensional frameworks \citep{Elia2020617}, while the latter include instruments to assess tangible (e.g., technological) and intangible constructs (e.g., skills or capabilities) that are essential to extracting the potential value of data \citep{Wu2022707}. In particular, capabilities associated with agents and technologies can be described in terms of \emph{maturity}, depending on the data-driven strategy. Big data maturity models are widely used by organisations to assess their readiness for adopting and implementing big data-driven strategies
\citep{Al-Sai_2022,Vesset_2015,Halper_Krishnan_2013, Corallo_Gervasi_2023}.  

The extraction of value from big data, often referred to as \emph{potential value}, is generally realised as an estimate of the expected value that an initiative could generate \citep{Ashton2007,Ishwarappa_Anuradha_2015, YlijokiPorras2016}. Specifically, the potential value hidden in (big) data is transformed in the transition from one stage of the DIKW model to another until it becomes visible through the measurement of business performance in the form of business value. In this paper, we focus on knowledge and the sources of uncertainty that could undermine its transition to wisdom. 

Actually, most data-driven strategy implementations fail to deliver the estimated value \citep{Reggio_Astesiano_2020}, and the causal factors that can explain the failure of data-driven strategies are not contextual and isolated but systemic \citep{Gervasi_2023_(1)}. These failures (misalignment between the estimated and generated business value) may come from several factors that characterise data-driven strategies, such as the randomness of the project life cycle, the role of different human or artificial agents, the uncertainty of the results, the multiple characteristics of (big) data, the obsolescence of technologies as a function of time, and the new types of value to be generated \citep{Gervasi_2023_(1)}.

This discrepancy between the expected and observed values suggests a need for a deeper investigation to better understand the relation between them. In fact, only a few organisations have implemented a structured value measurement system to rigorously quantify the return on investment in their data-driven strategy \citep{Grover2018}.

Finally, a key point that characterises data-driven strategies is the role of AI. The limited interpretability of intermediate outcomes during such processes also complicates the adaptation of decision strategies during the initiative. This is now opening up new questions regarding Explainable AI (XAI; see, e.g., \citet{Gunning2019}), trustworthy AI \citep{Floridi2019}, as well as human-centred and general-purpose AI.

\subsection{Scope of the work}
\label{subsec: scope of the work}

The present work addresses the need for useful representations of uncertainty about knowledge in data-driven strategies. A major source of uncertainty regards the \emph{observability} of data. Specifically, big data are not directly observable; rather, they require a proxy agent (e.g., automated tools) for processing. The effects of this lack of observability on the transition from information to knowledge and wisdom levels in the DIKW hierarchy become particularly evident in analyses based on deep learning, as they are characterised by \textit{black box} approaches that limit the interpretability of extracted information (e.g., selected features). Addressing this issue prompts the search for new approaches that enhance explainability in the adoption of tools relying on AI \citep{Gunning2019}. In this context, our goal is to provide a definition of explainability suitable for the type of uncertainty arising in data-driven strategies, as this uncertainty can undermine the assessment or measurement of the generated value.

For this purpose, we adopt a structural approach for modelling and representing some of the aforementioned notions. Our focus is the (in-)equivalence of relational systems that abstract the notions of accessible knowledge as a resource supporting decision-making and the possibility to transfer (i.e., explain) such knowledge. The objective does not limit itself to knowledge mediated by artificial agents nor to data-driven strategies; in principle, this abstraction level fosters a wider application of the identified properties to different scenarios characterised by ``non-classical'' (i.e., non-probabilistic) uncertainty. Indeed, we take advantage of formal analogies with other scenarios characterised by ambiguity, such as Ellsberg's urn models \citep{Ellsberg1961}, and explore inequivalent descriptions arising from different data observation levels in measurement settings, such as in Wigner's friend experiment \citep{Wigner1995,Frauchiger2018}. Starting with a dimensional definition of value, we identify order-theoretic structures to evaluate the (lack of) explainability in terms of the update of a knowledge representation. 

The relational structure is expressed by endowing dimensional frameworks with enough structure to recognise inconsistencies and ambiguities. In particular, we pay special attention to potential obstructions to the \emph{update} of a knowledge representation in terms of explainability. This can support organisational processes where new evidence and intermediate evaluations during the initiative continuously lead to changing objectives and adopting an agile management methodology.  

This high-level framework allows us to identify order-theoretic conditions that enable the description of inconsistencies or bounded resources in the update of a knowledge representation. Order-theoretic notions underlie several decision-making methodologies \citep{Tversky1969,Jamison1973,Greco2010}, which makes the assessment of such conditions scalable in multiple contexts. In the final part of this work, we discuss practical implications and comment on the link with consolidated methodologies to measure abstract constructs (e.g., Structural Equation Modelling, or SEM). This link can prompt future developments in the design of appropriate assessment tools and prevent ambiguities that may undermine a proper assessment of (big) data-driven strategies \citep{YlijokiPorras2016}.

The rest of this paper is organised as follows: in Section \ref{sec: value uncertainty scenarios}, we provide a brief description of uncertainty scenarios that are relevant to our discussion, especially ambiguity, measurements mediated by other agents, and deviations from classical or rational behaviours in decision-making. 
We start presenting the basic notions of our formalism in Section \ref{sec: from dimensional frameworks to dimensional structures}, then we pay special attention to the representation of knowledge states as well as their updates and explainability in Section \ref{sec: model proposal}. The different layers of knowledge inequivalence within our framework, from logical to information-theoretic, are deepened in Section \ref{sec: properties of knowledge frames and their relations to data-driven strategies}. In Section \ref{sec: modelling in data-agent interactions}, we formalise the mapping between our formalism and the uncertainty scenarios. In Section \ref{sec: scope of the work}, we discuss our approach in relation to specific forms of uncertainty in measurement models and inequivalence within multi-dimensional frameworks for data-driven strategies. Conclusions and future work are summed up in Section \ref{sec: conclusion}.

\section{Preliminaries on uncertainty scenarios}
\label{sec: value uncertainty scenarios}

To better clarify the notion of knowledge within the DIKW hierarchy, we begin by reporting the interpretation of value associated with each stage in Table \ref{tab: DIKW}  \citep{Zeleny_Only_1987,Lamba2015}.
\begin{table}[h!]
\begin{tabular}{|l|ll|l|}
\hline
\multicolumn{1}{|c|}{\textbf{Stage}} & \multicolumn{2}{c|}{\textbf{Taxonomy of knowledge}} & \multicolumn{1}{c|}{\textbf{Value}} \\ \hline
Data & \multicolumn{1}{l|}{Muddling through} & Know-Nothing & Nothing \\ \hline
Information & \multicolumn{1}{l|}{Efficiency (measurement+search)} & Know-How & Reveals relationships \\ \hline
Knowledge & \multicolumn{1}{l|}{Effectiveness (decision making)} & Know-What & Reveals pattern \\ \hline
Wisdom & \multicolumn{1}{l|}{Explicability (judgment)} & Know-Why & Reveals action or purpose \\ \hline
\end{tabular}
\caption{DIKW model \citep{Ackoff1989} associated with the \textit{Taxonomy of knowledge} \citep[Tab. 1]{Zeleny_Only_1987} managerial and descriptive interpretation, and the corresponding value associated with each stage \citep[Tab. 2]{Lamba2015}}
\label{tab: DIKW}
\end{table}

With reference to this chart, the patterns supporting decision-making should be interpreted or explained in line with a dimensional measurement of the value generated through actions. The occurrence of multiple dimensional assessments may lead to incompatible measurement settings. In the following paragraph, we illustrate relevant sources of indeterminacy that constitute the basis for our model construction.

\subsection{Uncertainty types in data-driven strategies}
\label{subsec: sources of uncertainty in data-driven strategies}

A study by \citet{Manyika} predicted that companies using big data in their product and service innovation processes save up to $20-30\%$ on product development costs and achieve faster time-to-market cycles by $50-60\%$. Similarly, in the public sector, the estimated cost reduction for administrative activities using big data was $15-20\%$, resulting in a projected generated value of 150 - 300 billion euros \citep{Cavanillas2016}. 

Despite their potential, there is a high failure rate for big data projects, reported to be as high as $85\%$ in \citet{Reggio_Astesiano_2020}. 
\citet{Montequin} conducted a study identifying and analysing $26$ \textit{failure causes} of Information and Communication Technologies (ICT) projects, as well as $19$ \textit{success factors}, using targeted questionnaires.
Among the former, the most common causes are incorrect or incomplete definitions of requirements, their continuous change even in the advanced stages of the project, and inaccurate estimations of costs and time. On the other hand, a clear vision of the project objectives and an accurate estimation of feasibility and costs appear among the major success factors.

Here, we concentrate on two types of uncertainty that can have a major impact on the assessment of data-driven strategies. The first type is \emph{ambiguity}, namely, the lack of a known or estimated probability distribution over the event space. Ambiguity refers to ``\textit{unknown unknowns}'', which have been extensively studied in decision- and strategy-making contexts; see, e.g., \citep{Rindova2020}. Among the relevant examples of decision frameworks with ambiguity, Ellsberg's urn models are recognised as a paradigm where classical approaches based on maximum expected utility fail to properly represent the preference patterns that are empirically observed \citep{Ellsberg1961,Sozzo2020}. Ambiguity relates to data-driven strategies mainly due to the lack of accurate or realistic estimates of the raw data value, but also to the mismatch between expected and observed success rates in data-driven strategies. 

The definition itself of value is uncertain, as the notion of value is an abstract construct; hence, the measurand is subject to metrological uncertainty. A potential manifestation of such a type of uncertainty is the inconsistency among the multi-dimensional frameworks that arose in the scientific literature to encompass the different realisations of value \citep{YlijokiPorras2016}. To reproduce this metrological uncertainty in our formalism, we link to the Wigner’s Friend thought experiment \citep{Wigner1995}, which focuses on the implications of non-classical (quantum) measurements. Even though this second scenario originated from the analysis of physical experimental settings, it provides a general basis to illustrate \emph{observer dependence} in evaluations and measurements, which is the second type of uncertainty we focus on in data-driven strategies. In fact, an explicit connection between the Wigner's Friend experiment and frameworks to reason about knowledge \citep{Halpern2017} was identified \citep{Frauchiger2018} and further extended \citep{Nurgalieva2018}. More generally, quantum-inspired methods proved useful in modelling the ambiguity in Ellsberg's models \citep{Aerts2018} and representing deviations from classical behaviours in cognition \citep{Sozzo2017}, decision-making \citep{Sozzo2020}, logic and operational theories \citep{Abramsky2014,Abramsky2017}, and social sciences \citep{Cervantes2019}.

We introduce these scenarios for ambiguity and observer dependence in evaluations, as we specify the two topics in our framework jointly in Section \ref{sec: modelling in data-agent interactions}.

\subsection{Ambiguity: urn models}
\label{subsec: Urn Models}

Urn models encompass a large class of measurement designs that realise various forms of uncertainty, including non-probabilistic ones. A pivotal example is the three-colour urn model introduced by Ellsberg  \citep[pp. 653-654]{Ellsberg1961}, 
which we now briefly describe. 

Let us consider an urn containing $90$ coloured balls, where $30$ of them are red and the remaining $60$ are yellow or black. The proportion of yellow and black balls is unknown to the decision-maker, who faces cost-free betting alternatives:
\begin{enumerate}
    \item $\pi_{0,a}$: get $100$ if a red ball is drawn from the urn; 
    \item $\pi_{0,b}$: get $100$ if a black ball is drawn from the urn; 
    \item $\pi_{1,a}$: get $100$ if a red \emph{or a yellow} ball is drawn from the urn; 
    \item $\pi_{1,b}$: get $100$ if a black \emph{or a yellow} ball is drawn from the urn.
\end{enumerate}
We introduce the symbol $\prec_{\mathrm{d}}$ to denote the preference relation of a decision-maker between the above-mentioned alternatives. Notably, findings have revealed preferences 
\begin{equation}
    \pi_{0,b} \prec_{\mathrm{d}} \pi_{0,a} \quad\text{and}\quad \pi_{1,a} \prec_{\mathrm{d}} \pi_{1,b}
\label{eq: violation Sure-Thing principle}
\end{equation}
which contradicts the subjective expected utility theory \citep{Ellsberg1961}. Specifically, (\ref{eq: violation Sure-Thing principle}) violates the \emph{Sure-Thing Principle}, one of Savage's postulates that can be included in the expected utility theory as a form of monotonicity between preferences and their expected utility. We refer to \cite{Aerts2018} for more details on this topic and for a framework that uses quantum structures to formalise this form of ambiguity in decision-making.

This model is discussed in the context of the framework presented in this work in Subsection \ref{subsec: connecting ambiguity and data-agent interactions}.

\subsection{Lack of observability: Wigner's Friend}
\label{subsec: Wigner's Friend}

Wigner's Friend is a central thought experiment in quantum physics, originally conceived by Wigner (see, e.g., \citep{Wigner1995}) to highlight the crucial role of observers in quantum measurements. We summarise the key aspects of this thought experiment that are relevant to our scope, directing readers to \citet{Frauchiger2018} and references therein for a more comprehensive discussion. 

We delve into the Wigner's Friend scenario based on the formal analogy between the notion of observability of data in our framework and the role of measurements and observers in quantum physics, both leading to an update of states. Wigner's experiment envisages two laboratories and two observers. The first observer, known as Wigner's Friend, is situated in a laboratory along with a measurement setup. Wigner himself serves as a ``super-observer'' outside the first laboratory and can perform measurements on it. Wigner's Friend performs the measurement on a physical system (a \emph{spin}), whose possible outcomes are denoted as $|+1\rangle$ and $|-1\rangle$. 

The question arises when Wigner's Friend actually observes the outcome of the measurement, while Wigner only knows that the measurement has been performed by his friend, but he has not measured it. For Wigner's Friend, the state is $|+1\rangle$ or $|-1\rangle$, depending on the outcome; on the other hand, Wigner attributes to the combined system in the lab (including the experimental setting and its friend) a superposition of two states, each one associated with the composition (product) of the observed outcome and the state of the friend: the measured system and the friend are \emph{entangled} for Wigner. Then, we have two different perspectives associated with the two observers, which leads to ambiguity about the system's state, namely, inconsistency associated with measurements that can be experimentally tested. 

\subsection{Evidence of non-classical behaviour in cognition and self-assessment}
\label{subsec: evidence of non-classical behaviour in cognition and self-assessment}

The aforementioned scenarios are practical examples that highlight the deviation from classically expected behaviour. 
Similar phenomena are also observed in realistic scenarios, where assumptions to be tested are stated through logical expressions that make use of classical disjunction, conjunction, and negation. These connectives allow for defining events and measuring them using classical approaches rooted in Kolmogorov probability axioms and Bayesian criteria for conditioning and updating probabilities. Measurement tools to assess cognitive constructs, such as questionnaires, enable the estimation of these probabilities from frequencies and the study of empirical correlations. Assumptions regarding abstract constructs and relations among them can be estimated through factor analysis and SEM \citep{Henseler2015,Henseler2016,Carpita2017,Ingusci2023}. In our scope, the constructs of interest are linked to maturity. 

Assessing an organisation's maturity in terms of capabilities, attitudes, and resources is a key point in defining and implementing data-driven strategies \citep{vandeWetering2019_b}. \citet{Corallo_Gervasi_2023} analysed the main maturity models according to the three groups of attributes proposed by \citet{Mettler_2009}. The first group refers to the general attributes, which are inherent in the basic information about models. The second group involves design attributes, which model the structure in terms of evaluation, scope, dimensions, maturity levels, design focus, and evaluation method. Finally, there are attributes related to model application, scope of use (e.g., descriptive, comparative, prescriptive), method of application (e.g., self-assessment, external assessments), and potential availability of supporting material. 

Although the models developed and adopted by organisations and analysed in the literature follow specific standards \citep{Gökalp_2021, de_Bruin2005}, the interpretability of the results is complex, and maturity models are often only descriptive or comparative but rarely prescriptive. 
As highlighted in \cite{Corallo_Gervasi_2023}, maturity models can have different designs in terms of the number of dimensions and scoring method. Thus, the measured maturity depends on these factors, which affect the measurement's reliability. In addition, it must be considered that the responses collected from respondents may be influenced by biases associated with sample selection. Therefore, the investigation of potential sources of uncertainty in the design of such assessment tools is essential for conducting proper analysis and getting useful insights from the acquired information about capabilities. 

Non-classicality arises, for example, when we find a misalignment between syntactic expressions based on classical logic, probability axioms, and empirical frequencies. Incompatibility may correspond to the lack of monotonicity $p(A\wedge B)\leq \min\{p(A),p(B)\}$ for events $A,B$ weighted by the probability $p(\cdot)$. This type of (conjunction) fallacy has been observed in questionnaires \citep{Tentori2004} and web searches \citep{Sozzo2017}. 
Even in this case, the explanation of such phenomena can benefit from the Hilbert space representation of quantum mechanics \citep{Sozzo2020}. In Section \ref{sec: scope of the work}, we discuss the implications of our formalism in light of the design of measurement tools that can properly account for non-classical uncertainty in maturity assessment.

Finally, we mention that other deviations from rational (Bayesian) behaviour regard contextuality, namely, the dependence on an observed property on the whole experimental setting, which includes other simultaneously measured properties. This characterising aspect of quantum phenomena \citep{Abramsky2014,Abramsky2017} extends to psychological measurements \citep{Dzhafarov2016}. Empirical demonstrations of contextuality in psychological assessments have been conducted based on the verification (or violations) of conditions implied by a classical model, namely, Bell-type and CHSH inequalities \citep{Cervantes2019}.  

\section{From Dimensional frameworks to Dimensional structures}
\label{sec: from dimensional frameworks to dimensional structures}

Building upon the discussion in the previous sections, we now address the assessment of knowledge value within a decision process for a data-driven initiative. For this purpose, we start with a brief summary of dimensional frameworks in the scientific literature. 

\subsection{State of the art}
\label{subsec: from dimensional frameworks to dimensional structures}
Dimensional frameworks are often used to characterise value, enabling its observation and measurement. \citet{Grover2018} distinguished between the \textit{functional value}, e.g., market share and financial return, and the \textit{symbolic value}, which can be identified in the impact on brand and reputation, leading to a positive image as a result of big data analytics (BDA) investment (\textit{signalling effect} or \textit{herding effect}).
\citet{Gunther2017} conducted a literature review on how organisations realise value from big data through ``\textit{paths to value}''. In \citep[Sect. 3.1.2]{Gunther2017}, the authors also discussed the current debate regarding the relation between algorithmic and human-based intelligence; we pay attention to this topic in Subsection \ref{subsec: self-reference}. \citet{FossoWamba2015} analysed the five criteria (dimensions) discussed by \citet{Manyika} and interpreted them as a different type of generated value. 

\citet{Gregor2006} conducted a large-scale survey involving more than a thousand organisations. The collected data also include information regarding ICTs in the organisation, the environment, the structure and management practices, and the perceived business value of the use of specific technologies. Finally, factor analysis was used to investigate significant constructs within the high-dimensional survey response data. This approach is of interest in the assessment of the goodness of the chosen dimensions, which can provide insights into possible drivers. The assumptions underlying the choice of model by \citet{Gregor2006} are based on subjective assessments by the authors, which they acknowledge as a limitation of this approach. Secondly, the results of these analyses contribute to a change in the companies themselves: in the Authors' words, ``[\textit{a}] \textit{number of these outcomes equip the firm for further change in a step-by-step process of mutual causation}'', which shows how the representation of value can evolve over time according to organisational and contextual changes, and vice versa.

\citet{Elia2020617} carried out a systematic literature review to investigate the representations of value; furthermore, they proposed a framework to identify the various types of value, defining $11$ \textit{value directions} and grouping them into dimensions. For this purpose, the authors started with the four dimensions of value in \cite{Gregor2006} and considered $22$ types of \textit{information technology benefits}. Although the framework presented by \citet{Elia2020617} takes its cue from Gregor and co-authors' framework and shares four common dimensions, the two models are distinct and do not lead to the same conclusions: changes prompted by internal (e.g., organisational) or external (e.g., contextual) factors discussed above may require not only the inclusion of a new value dimension but also a different value structure for existing ones. 

\subsection{Dimensional frameworks as state transitions}
\label{subsec: dimensional frameworks as state transitions}

We can consider the shift from the model proposed by \citet{Gregor2006} to the one proposed by \citet{Elia2020617} as an update of the dimensional architecture representing value. Another case where we can recognise the update of such a dimensional architecture can be found in \cite{Maçada_etal2012}. Starting with the models in \cite{Gregor2006} (4 \textit{supra-dimensions}) and \cite{Weill_Broadbent_1998} (1 \textit{supra-dimension}), \citet{Maçada_etal2012} identified a new model that confirms the four \textit{supra-dimensions} in \cite{Gregor2006} but permutes the \textit{sub-dimensions}. This aspect is worth considering because it stresses a subjective component in dimensional definitions, which are representative of a particular view, not only in terms of granularity but also in terms of classification.  

We start introducing our conceptual formalism to encompass this type of update by specifying the role of dimensional frameworks within an assessment process. 
Choosing a set of evaluation stages within the time span of the process, referred to as \emph{process states}, we focus on classes of transitions between them to highlight the relational aspects of value. For each pair of states $\psi_{1}$ and $\psi_{2}$, we associate a labelled transition between them and denote it as $\psi_{1}\overset{{\scriptstyle \tau}}{\rightarrow} \psi_{2}$. In this way, given a state $\psi_{1}$, the class of all the possible transitions $\tau$ originating from $\psi_{1}$ defines the possible inferences that an agent can make starting from $\psi_{1}$.

Now we include in our model the occurrence of multiple dimensions that guide the decision process and its evaluation. The minimal structure that we assume to support decision-making criteria is an order relation. So we provide the following: 
\begin{definition}
\label{def: value frames}
Let $\mathcal{V}$ be a set of partially ordered sets (\emph{posets}; see, e.g., 
\cite{Davey2002}). We label each element of $\mathcal{V}$ through a totally ordered set $\mathcal{I}$, so we can express 
\begin{equation}
    \mathcal{V}:=\{(V_{i},\preceq_{i}):\,i\in\mathcal{I}\}
\end{equation}
where $\preceq_{i}$ is a reflexive, antisymmetric, and transitive relation on $V_{i}$ for each $i\in\mathcal{I}$. 

Let $\mathcal{J}\subseteq \mathcal{I}$. We consider the categorial product of the latent dimensions $V_{i}$, $i\in \mathcal{J}$, which is the well-known Cartesian product for the category of sets ($\mathbf{Set}$): 
\begin{equation}
    \varrho_{\mathcal{J}} := \prod_{j\in\mathcal{J}} V_{j}
    \label{eq: product of posets} 
\end{equation}  
where the order of factors is derived from the order in $\mathcal{I}$. The corresponding \emph{value frame} is then defined as the disjoint union of such products for all non-empty subsets of $\mathcal{I}$:
\begin{equation}
    \varrho := \coprod_{\emptyset\subset\mathcal{J}\subseteq \mathcal{I}} 
    \varrho_{\mathcal{J}}
    \label{eq: exhaustive model of value}
\end{equation}
where $\coprod$ denotes the disjoint union (or coproduct) of sets.
\end{definition}

For the sake of concreteness, we present an example with relevant dimensions for data-driven strategies by recalling the framework provided by \citet{Polimeno2019}.

\begin{example}
\label{exa: disjoint union} 

The definition of the dimensions in \cite[Tab. 10]{Polimeno2019} refers to big data. In this example, we report two selected dimensions $V_{1} = \{m_1,m_2,m_3,m_4,m_5\}$ and $V_{2} = \{b_1,b_2,b_3\}$ where the interpretation of these labels is detailed as follows \citep[Tab. 5]{Polimeno2019}: 
\begin{itemize} 
\item $V_{1}$: \textit{Strategic value}
\begin{enumerate} 
\item $m_{1}$: ``\textit{New competitive advantage}''
\item $m_{2}$: ``\textit{Alignment between IT and business strategy}''
\item $m_{3}$: ``\textit{Quicker response to change}''
\item $m_{4}$: ``\textit{More effective customer relationships}'' 
\item $m_{5}$: ``\textit{Better products and services}''
\end{enumerate}
\item $V_{2}$: \textit{Transformational value}
\begin{enumerate}
\item $b_{1}$: ``\small{\textit{Reinforcement of organizational capabilities}}''
\item $b_{2}$: ``\small{\textit{Innovation in business models}}'' 
\item $b_{3}$: ``\small{\textit{Efficiency in organizational structure and processes}}''.
\end{enumerate}
\end{itemize} 
The authors drew these dimensions from \citet{Gregor2006}, who used factor analysis to investigate survey responses and associate $22$ benefit items with $4$ dimensions. In particular, ``\textit{Expanding organizational capabilities}'' has factor loadings of $0.49$ and $0.44$ along the Transformational and Strategic benefits, respectively. 

Starting with dimensions $V_{1},\dots,V_{n}$, we can construct a new representation by introducing $\mathcal{I}:=\wp(\{1,\dots,n\})\setminus\{\emptyset\}$, the set of non-empty subsets of $\{1,\dots,n\}$, and dimensions $W_{\mathcal{S}}$, $\emptyset\subset\mathcal{S}\subseteq \{1,\dots,n\}$. The elements of $W_{\mathcal{S}}$ are constructed from the values of benefit items in $\bigcap_{j\in\mathcal{S}} V_{j}$ when this intersection is not empty.  This allows distinguishing the membership of the ``\textit{Expanding organizational capabilities}'' item in different dimensions based on distinct interpretations. 
\end{example}

If one relies solely on the magnitude of factor loadings as a membership criterion to specify the dimensional framework, then the narrow difference between the loadings in the previous example might suggest an association of the benefit ``\textit{Expanding organizational capabilities}'' with both the Transformational and Strategic and dimensions. 
This points out the need to assess the identity of variables among distinct dimensions. Each individual study can adopt other quantitative criteria to confirm the discriminant validity of the constructs and dimensions in a structural model \citep{Henseler2015}. However, the combination of multiple studies should assess whether the measurement models are compatible. This fundamental requirement is made explicit for individual studies involving a multi-group analysis \citep{Henseler2016,Ingusci2023}, where the configural invariance is the first step to be checked to make this comparison meaningful \citep[pp. 413-414]{Henseler2016}. 

The comparability of distinct frameworks leads to additional issues regarding the identity of the dimensions investigated in the individual studies, even in terms of the interpretation of the indicators within the different measurement settings. Products and disjoint unions in (\ref{eq: exhaustive model of value}) can represent such deviations from unidimensionality in the sense of multiple latent dimensions related to a group of indicators. In data-driven strategies, such a distinction acquires more relevance due to the intrinsic ambiguity in the definition of (big) data-characterising features \citep{YlijokiPorras2016}.  

However, value frames in Definition \ref{def: value frames} do not take into proper account the lack of knowledge regarding the whole class of dimensions and potential interdependencies among them. Therefore, along with products and disjoint unions, we should also include meta-reasoning to highlight inconsistencies in the comparison of dimensional structures. For example, we can encode knowledge about the framework dimensionality through the inclusion of a representative of the state itself, e.g., a dimension $V_{\dim}:=\wp(\mathbb{N}_{0})$ that addresses the number of potential dimensions of the knowledge state. The choice of $\kappa\in\varrho_{\mathcal{J}}$ lets us represent uncertainty about the dimensionality ($\dim\in\mathcal{J}$ and $\#\kappa(\dim)>1$, where $\# S$ is the cardinality of a set $S$), as well as the lack of specification ($\dim\notin \mathcal{J}$). Furthermore, inconsistency arises when $\# \mathcal{J}\notin \kappa(\dim)$.

Next, we extend the value frame $\varrho$ by endowing it with relational structures and conditions to formalise the types of uncertainty mentioned in Section \ref{subsec: sources of uncertainty in data-driven strategies}.

\section{Model proposal: operational aspects of knowledge representations}
\label{sec: model proposal}

In this section, we explore the conditions that make knowledge transferable between different representations and, hence, reusable.

\subsection{Comparison, composition, and the role of observers: inner states}
\label{subsec: Comparison and composition}

The assumption that each dimension $V_{i}$ is endowed with a partial order $\preceq_{i}$, i.e., a reflexive, symmetric, and transitive, but not necessarily total relation, is in line with well-established methods to formalise concept analysis and knowledge structures \citep{Doignon2012}. A stronger assumption that can be considered is the existence of an associative operation $\oplus_{i}$ for each latent dimension $V_{i}$ that is \emph{idempotent}, i.e., $x\oplus_{i} x = x$ for all $x\in V_{i}$ and $i\in \mathcal{I}$. This operation defines an order relation as follows:
\begin{equation}
    \forall a,b \in V_{i}:\quad a\preceq_{i} b \Leftrightarrow a \oplus_{i} b = b. 
    \label{eq: tropical addition to poset}
\end{equation}
Such an operation, which combines the partial order $\preceq_{i}$ and the notion of composition, is essential for the subsequent analysis of knowledge representations in the context of data-driven strategies. To establish an algebraic structure for comparing and composing different knowledge states, we endow the value frame $\varrho$ with an order relation $\preceq$ as follows: we express the content of each individual element $\kappa\in\varrho$ as a partial function from $\mathcal{I}$ to $\coprod_{i\in\mathcal{I}} V_{i}$ by defining, for a given $\mathcal{J}\subseteq \mathcal{I}$, the map  
    \begin{equation}
        \kappa:\, \mathcal{J} \longrightarrow \coprod_{i\in\mathcal{I}} V_{i},\quad j\mapsto \kappa(j) \in V_{j}.
        \label{eq: inner state example, indicator function}
    \end{equation}
For each $\mathcal{H}\subseteq\mathcal{J}\subseteq \mathcal{I}$, $\pi_{\mathcal{H}}$ denotes the projection of $\varrho_{\mathcal{J}}$ onto $\prod_{h\in\mathcal{H}}V_{h}$, and we explicitly write $\pi_{\mathcal{J},i}$ to refer to the canonical projection onto $V_{i}$ for each $i\in\mathcal{J}$. Let us denote the domain of such a partial function as $\underline{\kappa}$. For all $\kappa_{1},\kappa_{2}$ with $\underline{\kappa_{1}} = \underline{\kappa_{2}}$, we consider $\kappa_{1}\preceq \kappa_{2}$ if 
    \begin{equation}
        \forall i\in \underline{\kappa_{1}}:\,\pi_{\underline{\kappa_{1}},i}(\kappa_{1}) \preceq_{i} \pi_{\underline{\kappa_{2}},i}(\kappa_{2}).
        \label{eq: componentwise dominance}
    \end{equation}
This definition is in line with the notion of product category, with particular reference to posetal categories in the present discussion. Then, we extend this order by taking into account the domain of the elements of $\varrho$: 
    \begin{equation}
        \kappa_{1}\preceq \kappa_{2} \Leftrightarrow  \underline{\kappa_{1}} \subseteq \underline{\kappa_{2}}\quad \text{and} \quad \kappa_{1} \preceq \kappa_{2}|_{\underline{\kappa_{1}}}
        \label{eq: domain extension}
    \end{equation}
where $\kappa_{2}|_{\underline{\kappa_{1}}}$ denotes the restriction of $\kappa_{2}$ to the domain  $\underline{\kappa_{1}}$ of $\kappa_{1}$. We denote the resulting poset $(\varrho,\preceq)$ as $\varrho^{\star}$. Now we can state the following:
\begin{definition}
\label{def: domain and codomain of inner state and sections} 
An \emph{inner state} $\kappa$ is an element of the poset $\varrho^{\star}$. With a slight abuse of notation, we can equivalently express each individual inner state $\kappa$ as a partial function with domain $\underline{\kappa}$, in line with (\ref{eq: inner state example, indicator function}). In this way, we can compare inner states by domain extension (\ref{eq: domain extension}) and componentwise ordering (\ref{eq: componentwise dominance}).
\end{definition}

The statement $a \preceq_{i} b$ can be interpreted as follows: let us consider any two agents $\mathtt{A},\mathtt{B}$ with inner states $\kappa_{\mathtt{A}},\kappa_{\mathtt{B}}$ respectively, such that $i\in \underline{\kappa_{\mathtt{A}}}\cap \underline{\kappa_{\mathtt{B}}}$, $a=\pi_{\underline{\kappa_{\mathtt{A}}},i}(\kappa_{\mathtt{A}})$, and $b=\pi_{\underline{\kappa_{\mathtt{B}}},i}(\kappa_{\mathtt{B}})$. Then, all the knowledge value recognised by Agent $\mathtt{A}$ along the dimension $V_{i}$ is also recognised by Agent $\mathtt{B}$. In our perspective, the order relation represents the possible inferences that can be drawn by an agent using its knowledge resources. 

\begin{remark}
The focus on the domain of an inner state is of major relevance in defining awareness in the present formulation of data-driven strategies. Indeed, the statement $i\in\underline{\kappa}$ is interpreted as the assertion that the agent can evaluate its knowledge along the dimension $V_{i}$. If $\perp_{i}$ is the minimum element of $V_{i}$, assuming it exists, the statement $\kappa(i)=\perp_{i}$ means that the agent knows she does not know about the dimension $V_{i}$. On the contrary, $i\notin\underline{\kappa}$ means that the agent is unaware of the dimension $V_{i}$. 

This can be read in relation to the operator $K_{i}$ associated with Agent $i$ in modal logic; the combination of the notions that are used to model knowledge structures (in particular, Kripke structures) in the Wigner's Friend extended scenarios designed by \citet{Frauchiger2018} is discussed in \citep{Nurgalieva2018}.
\end{remark}

The order (\ref{eq: domain extension}) extends (\ref{eq: componentwise dominance}) by relaxing a comparability condition, from $\underline{\kappa_{1}}=\underline{\kappa_{2}}$ to $\underline{\kappa_{1}}\subseteq \underline{\kappa_{2}}$. An analogous distinction was pointed out in \citep[Sect. 3]{Angelelli2017} to examine inequivalent representations in a statistical physical context. In the scope of this work, we note the following:
\begin{remark}
\label{rem: upper and lower poset extensions}
The domain extension (\ref{eq: domain extension}) establishes a connection between different dimensional contexts $\varrho_{\mathcal{J}}$. This generates the order compatibility:
\begin{equation}
\kappa_{1}|_{\underline{\kappa_{1}}\cap \underline{\kappa_{2}}}\preceq \kappa_{2}|_{\underline{\kappa_{1}}\cap \underline{\kappa_{2}}}
\Rightarrow 
\kappa_{1}|_{\underline{\kappa_{1}}\cap \underline{\kappa_{2}}}\preceq \kappa_{2}
    \label{eq: order compatibility}
\end{equation} 
The order (\ref{eq: domain extension}) is a \emph{value-based} view, as it does not differentiate between different domains due to (\ref{eq: order compatibility}). Other orders can be associated with the set $\varrho$; in particular, we can compare two elements only if they have the same domain: 
\begin{eqnarray}
    \kappa_{1}\sqsubseteq \kappa_{2} \Leftrightarrow \underline{\kappa_{1}}=\underline{\kappa_{2}}\text{ and } \forall i \in \underline{\kappa_{1}}:\,\kappa_{1}(i) \preceq_{i} \kappa_{2}(i). 
    \label{eq: lower poset}
\end{eqnarray}
This second order is a \emph{domain-based} view, leveraging the representation of inner states as partial functions. In this sense, the comparability of two partial functions is not based only on the value of potential input dimensions but also on the fact that they are functions with the same domain. This defines a new poset, $\varrho_{\star}:=(\varrho,\sqsubseteq)$. Clearly, $\varrho^{\star}$ is an extension of $\varrho_{\star}$ since each pair $\kappa_{1}\sqsubseteq \kappa_{2}$ corresponds to a pair $\kappa_{1}\preceq \kappa_{2}$ in $\varrho^{\star}$. 

The two posets coincide only under the unidimensionality condition $\#\mathcal{I}=1$. In the multi-dimensional case, we refer to $\varrho_{\star}$ and $\varrho^{\star}$ as \emph{lower} and \emph{upper} posets, respectively. This terminology adapts the lower and upper probabilities (or belief and plausibility, respectively) that are used to model imprecise probability, e.g., in Dempster-Shafer theory \citep[Sect. 2.3-2.4]{Halpern2017} (also see \citet{Cuzzolin2020} for a geometric view of these notions).
\end{remark}

\subsection{Self-reference}
\label{subsec: self-reference}

The distinguishing role of the domain can be used as a basis to consider different orders starting from the same class of posets. In turn, this allows for modelling the partiality of the composition (\ref{eq: tropical addition to poset}). This objective fits the scope of this work, as the (lack of) composition of knowledge states can be linked to the (lack of) explainability. For example, given the knowledge representations $\kappa_{\mathtt{H}}$ and $\kappa_{\mathtt{A}}$ of a human agent $\mathtt{H}$ and an artificial agent $\mathtt{A}$, respectively, the composition $\kappa_{\mathtt{A}}\oplus \kappa_{\mathtt{H}}$ may not be feasible when the knowledge of $\mathtt{A}$'s inner state $\kappa_{\mathtt{A}}$ is partially accessible to $\mathtt{H}$.  

Looking at the scenarios described in Section \ref{sec: value uncertainty scenarios}, the type of uncertainty we want to describe primarily arises through meta-reasoning, which, in our context, involves the composition or comparison of inner states. In particular, the ambiguity in Ellsberg's urn model is not manifest when considering individual decision contexts (bets $(\pi_{0,a},\pi_{0,b})$ and $(\pi_{1,a},\pi_{1,b})$), but it emerges only when both of these contexts are taken into account and compared. Similarly, the Wigner's Friend phenomenon involves Wigner's reasoning about its friend's state, as elaborated in \citep{Frauchiger2018}. 

Then, we can use the two posets $\varrho_{\star}$ and $\varrho^{\star}$ defined in Subsection \ref{subsec: Comparison and composition} to represent this form of meta-reasoning. 

\begin{definition} 
\label{def: metastates} 
We extend $\varrho^{\star}$ by appending the poset $(\varrho,\sqsubseteq)$ as a new dimension and repeating the construction in (\ref{eq: componentwise dominance})-(\ref{eq: domain extension}). The corresponding poset  
\begin{equation}
    \mathrm{K}:=\left(\varrho\sqcup \varrho\sqcup (\varrho\sqcap \varrho), \preceq_{\mathrm{K}} \right)
    \label{eq: metastates}
\end{equation}
with the order $\preceq_{\mathrm{K}}$ induced by the aforementioned construction is referred to as the poset of \emph{lower meta-states}.
\end{definition}

In Remark \ref{rem: upper and lower poset extensions}, $\varrho_{\star}$ is introduced as a lower poset, namely, the inclusion $\sqsubseteq\subseteq \preceq$ between relations on $\varrho$ holds. In analogy to lower and upper posets, we can also consider a different definition of meta-states where $\varrho^{\star}$ plays the role of the lower poset, and we extend the order $\preceq$ in the new dimension. In particular, we can consider a linear extension $\leq$ of $\preceq$ by using a monotone function $h:\,\varrho^{\star}\longrightarrow \mathbb{R}$:  
\begin{equation}
    \kappa_{1} \preceq
    \kappa_{2} \Leftrightarrow h(\kappa_{1}) \leq h(\kappa_{2}). 
\label{eq: linear order from information measure}
\end{equation}
We use this representation in the following sections to provide an information-theoretic view of the state, where the extension $\preceq\subseteq\leq$ is derived from an information measure. This type of meta-reasoning can be expressed in a more general setting by introducing a second type of meta-states, which rely on a poset $V_{\text{exp}}$ that expresses knowledge about the relational structure in $\varrho^{\star}$. 

\begin{definition}
    \label{eq: nu-meta-states}
Let us consider a poset $V_{\text{exp}}$ and an order-preserving mapping $\nu:\, \varrho^{\star}\longrightarrow V_{\text{exp}}$. Then, we append $V_{\text{exp}}$ as a new dimension and obtain the poset 
\begin{equation}
\mathrm{K}^{(\nu)} := \left(\varrho\sqcup V_{\text{exp}}\sqcup (\varrho\sqcap V_{\text{exp}}), \preceq_{\nu} \right)
    \label{eq: representation-dependent poset of meta-states}
\end{equation}
where $\preceq_{\nu}$ in (\ref{eq: representation-dependent poset of meta-states}) is obtained through the construction in (\ref{eq: componentwise dominance})-(\ref{eq: domain extension}). 
The elements of the poset $\mathrm{K}^{(\nu)}$ are referred to as \emph{$\nu$-meta-states}.
\end{definition}

\subsection{Explainability as compositional existence}
\label{subsec: explainability as compositional existence}

Finally, we use the notions introduced above to provide a formal definition of explainability in our context. 

\begin{definition}
\label{def: explainability}
A lower meta-state is called \emph{diagonal} if it can be expressed as $(\kappa,\kappa)\in\mathrm{K}$ for some inner state $\kappa \in \varrho^{\star}$. 
The \emph{update} of $\kappa\in\varrho^{\star}$ by lower meta-states is the composition $\kappa\vee\psi$ with another state $\psi\in\varrho^{\star}$, when the supremum exists. Such an update is said to be \emph{explainable} when $(\kappa,\kappa)\vee(\psi,\psi)$ also exists in $\mathrm{K}$ and, hence, is diagonal. 

Similarly, a $\nu-$meta-state is diagonal if it has the form $(\kappa,\nu(\kappa))$  for some $\kappa\in\varrho^{\star}$. An explainable update by $\nu$-meta-states is a composition (supremum) of diagonal elements in $\mathrm{K}^{(\nu)}$ that is diagonal too.
\end{definition}

This definition specifies the possibility of extending an inner state and distinguishing extensions that change the knowledge base and, hence, are inconsistent with respect to the current dimensional setting. Note that the focus of explainability in this context is on knowledge updates, in agreement with the attention paid to (non-)reusable knowledge in data-driven strategies. When a combination $K\vee \kappa_{\mathtt{A}}$ is not feasible or is not compatible with the explainability accessible to $K$, the two inner states cannot be combined into a new state of knowledge. In turn, this may impede the reuse of the knowledge resource carried by another inner state in other contexts. An obstruction to the existence of such a state is the presence of multiple, non-equivalent value representations that an agent cannot directly discern. This is the situation we want to explore to describe interactions between agents in structural terms.

\begin{remark}
\label{rem: accessibility from projections}
The previous definition stresses the role of explainability in relation to accessible knowledge through the projections $\pi_{\underline{\kappa},i}$ (see Definition \ref{def: domain and codomain of inner state and sections}). According to Remark \ref{rem: upper and lower poset extensions}, the role of the domain of inner states in the definition of $\mathrm{K}$, e.g., through the lower poset $\varrho_{\star}$ in (\ref{eq: lower poset}), relates the existence of an explanation for a lower meta-state to the existence of the same set of projections $\pi_{\mathcal{J}}$, $\mathcal{J}\subseteq \underline{\kappa}$, for both $\kappa$ and $\kappa\vee \psi$. 
\end{remark}

In the following sections, we analyse the definitions provided above through inconsistencies that cannot be resolved through an explainable update of an inner state. Within a data-driven initiative, this allows assessing whether data and other agents that can observe them generate \emph{reusable} knowledge (an explainable update) or not. 

\section{Uncertainty and inequivalence in knowledge representations}
\label{sec: properties of knowledge frames and their relations to data-driven strategies}

The labelling induced by the disjoint product (\ref{eq: exhaustive model of value}) in this framework is analogous to Indexation-by-conditions in the Contextuality-by-Default approach in cognitive sciences, where variables are indexed by the context they are part of \cite{Dzhafarov2016}. To remove the dependence of such context, we can represent each element of $\kappa\in\varrho^{\star}$ as a tuple of pairs $(i,\kappa(i))$; then, we consider the equivalence relation defined by the projection onto the second coordinate, so $(i,\kappa(i))$ is identified with $(j,\kappa(j))$ if and only if $\kappa(i)=\kappa(j)$. 

Such labellings generate a connection between different posets, as elements of $V_{i}\cap V_{j}$, $i,j,\in\underline{\kappa}$, act as a linkage between the underlying posets $V_{i}$ and $V_{j}$. We can formalise this linkage by extending the focus from equal to corresponding elements through order-preserving mappings. This extension lets us consider the extent of order compatibility between different dimensions, since order-preserving mappings may not preserve compositions (suprema). Given the interpretation of explainable updates based on compositions (Definition \ref{def: explainability}), these order-preserved mappings can be used to realise inequivalent knowledge representations. 

Next, we present three instances of knowledge inequivalence from the logical, order-theoretic, and information-theoretic perspectives, respectively.

\subsection{Non-classicality from multiple potential implications}
\label{subsec: non-classicality from multiple potential implications}

The DIKW hierarchy mentioned in the Introduction distinguishes access to data from the decision-making stages, which can be summarised as a set of inferences to draw conclusions about the effect of actions based on evidence and empirical premises. 

The first layer where the inequivalence of knowledge representations can be addressed is the logical one, where inference is expressed as material implication. Specifically, uncertainty emerges as the occurrence of multiple inequivalent implications, which can be obtained in our context through the following construction.

\begin{example}
    \label{exa: lack of knowledge for complementation}
Let us consider a set-theoretic representation of a finite distributive lattice as an appropriate subset of the power set $\wp(\mathcal{S})$, which is always possible due to Birkhoff's representation theorem \citep[Sect. 5.12]{Davey2002}. It is well known that the implication $A\rightarrow \cdot$ defined by  $A \rightarrow Y:=Y\cup (A)^{\mathtt{C}}$ is the upper adjoint of the conjunction $A\cap \cdot$; namely, they are monotonic functions satisfying the relation 
\begin{equation}
    A \cap Y \subseteq Z \Leftrightarrow Y \subseteq A \rightarrow Z, \quad A,Y,Z,\subseteq \mathcal{S}.
    \label{eq: adjoint relation}
\end{equation}
where $\mathtt{C}$ denotes the set-theoretic complement with respect to $\mathcal{S}$. The implication is a fundamental logical connective to describe inference, and the adjointness condition is the basis for generalising classical logic to Heyting algebras and extended-order algebras in the context of fuzzy operators (see, e.g., \citep{DellaStella2012} and references therein).  

However, this construction presumes the knowledge of the whole set $\mathcal{S}$ to evaluate the complement $\mathtt{C}$. In particular, we can model partial knowledge on $\mathcal{S}$ by considering a class $\left\{\mathcal{S}_{u},\,u\in\{1,\dots,n\}\right\}$ of $n$ potential spaces that define as many implications $\rightarrow^{u}$, $u\in\{1,\dots,n\}$. 
\end{example}

\begin{remark}
\label{rem: orthomodular or not}
From the previous argument, we can see that the uncertainty about the base set $\mathcal{S}$ entails a deviation from classicality. Indeed, we have two alternatives for a finite poset $\mathrm{K}$. When $\mathrm{K}$ is a distributive lattice, the previous example shows that the lack of knowledge of the full set of dimensions generates multiple implications. Otherwise, the poset is not distributive, which is a main deviation from classical logic that is used to characterise quantum logics (in particular, by replacing distributivity with modularity; see, e.g., \citep{Harding1996}) and their extensions. In both alternatives, we infer a non-classical behaviour from bounded knowledge resources.
\end{remark}

From the logical layer, we can move our focus to the accessibility of potential inferences that can be drawn. In turn, this interpretation exploits the order structure of the dimensions and entails a second layer of inequivalence in knowledge representations. 

\subsection{Inequivalent representations from accessibility boundary}
\label{subsec: inequivalent representations from accessibility boundary}

Inequivalent descriptions of a statistical system are often a hint of its non-trivial characteristics. For instance, in \citep[Sect. 3]{Angelelli2017}, a limiting procedure changes the compositional structure of the statistical model, returning an instance of (\ref{eq: tropical addition to poset}) with connections to fuzzy sets. Such a limit accounts for a parameter configuration suited to exponential degenerations of energy levels. In view of our formalisation of the Wigner's Friend scenario, we investigate this aspect in our setting through the following example.

\begin{example}
\label{exa: non-equivalent representations}
Let us take a class of dimensions $\mathcal{V}$ and a distinguished dimension $V_{i}$ where the order relation $\preceq_{i}$ is partial. Furthermore, suppose that the supremum $a \vee_{i} b$ exists for all $a,b\in V_{i}$, which is an instance of (\ref{eq: tropical addition to poset}).  
Another way to encode the relation $\preceq_{i}$ is to introduce
\begin{equation}
    x\in V_{i}\mapsto \iota_{i}(x):= \{y\prec_{i} x:\,y\in V_{i}\}, \quad (V^{\star}_{i},\preceq^{\star}_{i}):=\left(\left\{\iota_{i}(x):\,x\in V_{i}\right\},\subseteq \right)
    \label{eq: micro-macro transformation}
\end{equation}
where $a\prec_{i} b$ means $a\preceq_{i} b$ and $a\neq b$. This representation of value is not based on a single valuation $x\in V_{i}$ but on the inferences that can be drawn through non-trivial processing of the available information encoded in $x$. 

The statement $a\preceq_{i} b$ implies that $\iota_{i}(a)\subseteq \iota_{i}(b)$, so $\iota$ is a strictly monotone mapping; in this sense, the order relation between compatible elements is preserved passing from $V_{i}$ to $V^{\star}_{i}$ (i.e., an order morphism). However, these two representations differ when the composition structure (\ref{eq: tropical addition to poset}) is considered: we can consider $\vee$ (the supremum operation) and $\cup$ as two operations satisfying (\ref{eq: tropical addition to poset}) for $V_{i}$ and $V^{\star}_{i}$, respectively. According to \cite[Prop. 5.1]{Angelelli2017}, we know that $(V_{i},\vee)$ is homomorphic to $(V^{\star}_{i},\cup)$ only if $(V_{i},\preceq_{i})$ is totally ordered. 
Since the order relation attributed to $V_{i}$ is not total, these different definitions of the dimension produce inequivalent results under composition. 
\end{example}

The final layer is the quantitative one, which relies on measures to evaluate the amount of knowledge encoded in a given representation. In the following subsection, we adapt the notion of information measure to our knowledge-based view.

\subsection{Uncertainty in information measures}
\label{subsec: uncertainty on information measures}

The lack of knowledge about the entire dimensional space ($\mathcal{S}$ in Subsection \ref{subsec: non-classicality from multiple potential implications}, unique maximal elements of $\iota(x)$ in Subsection \ref{subsec: inequivalent representations from accessibility boundary}) affects the quantification of the information content in knowledge representations. This lets us establish the analogy with the role of normalisation of subsystems discussed in \cite[Sect. 7]{Angelelli2017}. Consider a state $\kappa$ and the ordered set of non-negative reals $(\mathbb{R}_{\geq 0},\leq)$. Associate each dimension $V_{i}$, $i\in\underline{\kappa}$, with a weight $w_{\kappa}(i)$ for a given function $w_{\kappa}:\,\underline{\kappa}\rightarrow \mathbb{R}_{\geq 0}$. In particular, we can take a monotone function $\nu:\,\varrho^{\star}\longrightarrow \mathbb{R}_{\geq 0}$, which was also mentioned for the construction of $\nu$-meta-states in (\ref{eq: representation-dependent poset of meta-states}), and obtain $w_{\kappa}$ via $w_{\kappa}(i):=\nu\left(\kappa|_{\{i\}}\right)$. In this way, we take into account both the domain and the values of $\kappa$; by focusing on the dimension $\varrho_{\star}$ introduced in (\ref{eq: lower poset}), we can also associate each $\kappa\in\varrho_{\star}$ with a weight $w_{\kappa}(\varrho_{\star})\geq 0$, so that $\kappa_{1}\sqsubseteq \kappa_{2}$ implies $w_{\kappa_{1}}(\varrho_{\star})\leq w_{\kappa_{2}}(\varrho_{\star})$. On the other hand, the attribution $W_{\kappa}(i):=\nu\left(\bigvee V_{i}\right)$ only depends on the supremum $\bigvee V_{i}$ of $V_{i}$ and provides us with a definition of normalisation suited to our context, which only relies on the domain $\underline{\kappa}$.

Then, to identify uncertainty within information measures, we can use a representation where each $V_{i}\in\mathcal{V}$ is isomorphic to  $(\mathbb{R}_{\geq 0},\leq)$. In this setup, each $\kappa\in\varrho$ induces, after normalisation, a probability distribution whose support is $\underline{\kappa}$. We consider as an information measure $h:\,\varrho\longrightarrow \mathbb{R}_{\geq 0}$ the normalised Shannon entropy \citep[Sect. 3.11]{Halpern2017}
\begin{eqnarray}
    h(\kappa) & := & \frac{H(\kappa)}{H_{\max}(\underline{\kappa})}\nonumber \\ 
    & = & -\frac{1}{\ln\left(\#(\underline{\kappa})\right)}\cdot\sum_{t\in\underline{\kappa}}\frac{w_{t}}{\sum_{u\in\underline{\kappa}} w_{u}}\cdot \ln\left(\frac{w_{t}}{\sum_{u\in\underline{\kappa}} w_{u}}\right).
    \label{eq: normalised entropy}
\end{eqnarray}
This function is widely adopted to quantify the uncertainty (or, dually, the information and complexity) within a given distribution. The inclusion of the normalisation $\ln\left(\#(\underline{\kappa})\right)^{-1}$ derived from the maximum entropy achievable for a distribution with support $\underline{\kappa}$ takes into account the \emph{potential} probability assignments to the available dimensions. Such a dependence makes the normalised entropy undefined, and, when available, partial information about the support $\underline{\kappa}$ in terms of lower or upper approximations induces bounds for $\ln(\#(\kappa))^{-1}$. 

In this setting, each poset $\varrho_{\star}$ and $\varrho^{\star}$ entails order conditions between probability distributions. This kind of comparison also arises in information theory when one considers the \emph{Kullback-Leibler divergence} \citep[Ch. 3]{Halpern2017} to quantify the differences from the update of a probability distribution. Denoting as $p(\kappa_{1})$ and $p(\kappa_{2})$ the probability distributions associated with $\kappa_{1}$ and $\kappa_{2}$, respectively, the Kullback-Leibler divergence $D_{\mathrm{KL}}(p(\kappa_{2})||p(\kappa_{1}))$ from $p(\kappa_{1})$ to $p(\kappa_{2})$ can be evaluated only if $p(\kappa_{2})=0$ whenever $p(\kappa_{1})=0$. This assumption of absolute continuity 
formalises the constraint that the support of the distributions (interpreted as the set of elements with positive probability weight) does not increase. In our framework, we include the possibility to extend the support $\underline{\kappa_{1}}$ with new dimensions; on the other hand, this extension is evaluated differently in the two posets $\varrho_{\star}$ and $\varrho^{\star}$, which gives a criterion to discriminate explainable updates.

\section{Modelling in data-agent interactions}
\label{sec: modelling in data-agent interactions}

Now we examine the uncertainty scenarios described in Subsection \ref{sec: value uncertainty scenarios} within the proposed framework. 

\subsection{Ambiguity and data-agent interactions}
\label{subsec: connecting ambiguity and data-agent interactions}

\subsubsection{Preliminary discussion}
\label{subsubse: Ellsberg preliminary discussion}

Before defining the connection between Ellsberg's three-colour urn model and our formalism in Subsection \ref{subsubsec: ambiguity and non-explainable updates}, we identify some preliminary analogies to contextualise the decision-making problem in the scope of this work. The decision-maker is represented by a human agent $\mathtt{H}$, who has access to information regarding the value of the ``Red'' dimension; specifically, $\mathtt{H}$ can assess the risk associated with ``Red'', e.g., knowing its impact and probability. On the contrary, the information possessed by $\mathtt{H}$ about the remaining two colours, ``Black-Yellow'', only acknowledges their existence and their cumulative probability weight ($\frac{2}{3}$). 

A second agent, which we can associate with artificial intelligence (AI) with an inner state denoted as $\kappa_{\mathtt{A}}$, can get access to data to recognise the value of the ``Black-Yellow'' information dimension to a greater extent with respect to $\mathtt{H}$. In particular, there may be a latent factor in the data that lets $\mathtt{A}$ distinguish two ``Black'' and ``Yellow'' information dimensions. The human agent knows that $\mathtt{A}$ is able to recognise new value in the ``Black-Yellow'' dimension. 

Ellsberg's paradox corresponds to the misalignment between these information dimensions and knowledge dimensions, namely, the two decision contexts corresponding to the two lotteries $(\pi_{0,a},\pi_{0,b})$ and $(\pi_{1,a},\pi_{1,b})$. The existence of $\mathtt{A}$ allows the extraction of value from the ``Black-Yellow'' information dimension, but this cannot prompt a change of knowledge state for  $\mathtt{H}$ that is able to discern the value of the ``Black'' and the ``Yellow'' evaluations. 

\begin{figure}
\begin{center}
\usetikzlibrary{patterns}
\begin{tikzpicture}
\path [pattern=checkerboard,pattern color=black!10] (-2.5,0) rectangle (-1.5,2);
\path [pattern=checkerboard,pattern color=black!10] (-1,0) rectangle (3,2);
\node (v2) at (-2,1.5) {$\mathtt{Red}$};
\node at (0,1.5) {$\mathtt{Black}$};
\node (v5) at (2,1.5) {$\mathtt{Yellow}$};
\node (v3) at (1,0.5) {$\{\mathtt{Black},\mathtt{Yellow}\}$};
\node (v1) at (-1,4.5) [style={draw,shape=circle}] {$K_{\mathtt{H}}$};
\node (v4) at (1,3) [style={draw,shape=circle}] {$\kappa_{\mathtt{A}}$};
\draw [->] (v1) edge (v2);
\draw [->] (v1) edge (v3);
\draw [->] (v1) edge (v4);
\draw [->>] (v4) edge (v5);
\end{tikzpicture}
\caption{Diagrammatic representation of Ellsberg's three-colour model compared to the relation of human and artificial agents with data.}
\label{fig: Ellsberg three-colour model}
\end{center} 
\end{figure}
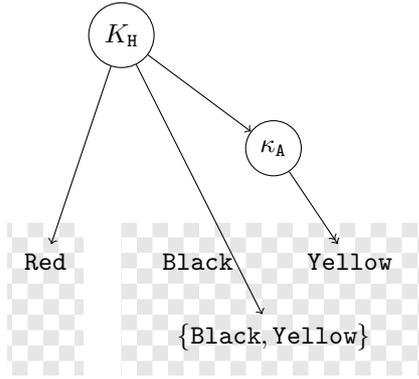

We provide a diagrammatic depiction of this phenomenon in Figure \ref{fig: Ellsberg three-colour model}. 
\vspace{0.5cm}

\subsubsection{Ambiguity and explainability of knowledge updates}
\label{subsubsec: ambiguity and non-explainable updates}

Now we provide a formal correspondence between Ellsberg's three-colour model and the present framework. We introduce the notation $\mathtt{B}^{(n)}$ for the Boolean algebra $\left(\wp(\{1,\dots,n\}),\cup,\cap,\cdot^{\mathtt{C}},\{1,\dots,n\},\emptyset\right)$ of the power set $\wp(\{1,\dots,n\})$. Consider two dimensions, $\mathtt{B}_{0}$ and $\mathtt{B}_{1}$, isomorphic to $\mathtt{B}^{(1)}$. These two objects abstract the two different observation/measurement settings, i.e., the two lotteries $(\pi_{0,a},\pi_{0,b})$ and $(\pi_{1,a},\pi_{1,b})$ in Ellsberg's model. The set $\mathrm{K}$ is in bijection with  
\begin{equation} 
\wp(\{\top^{(0)}\})\sqcup\wp(\{\top^{(1)}\})\sqcup\wp(\{\top^{(0)},\top^{(1)}\})
\end{equation} 
where we use the labellings $(0)$ and $(1)$ to distinguish the two lotteries as a result of the disjoint union. Then we focus on the explainability of the composition $\{\top^{(0)}\}\vee \{\top^{(1)}\}$ of $\{\top^{(0)}\}\in\mathtt{B}_{0}$ and $\{\top^{(1)}\}\in\mathtt{B}_{1}$, which represents an update where an agent becomes aware of a second decision scenario (lottery). We obtain   
\begin{equation} 
    \{\top^{(0)}\}\vee \{\top^{(1)}\}=\left(\{\top^{(0)}\},\{\top^{(1)}\}\right)\quad \text{in }\varrho^{\star}. 
    \label{eq: Ellsberg, composition in upper poset}
\end{equation}
On the other hand, this composition is not defined in $\varrho_{\star}$. 

An analogous result is obtained using $\nu$-meta-states as in Definition \ref{def: explainability}. Here we consider $\nu:\,\varrho^{\star}\longrightarrow \mathtt{B}^{(1)}$ with $\nu(\kappa)=\{1\}$ at $\underline{\kappa}=\{1,2\}$, and $\nu(\kappa)=\emptyset$ otherwise. From (\ref{eq: Ellsberg, composition in upper poset}), we find 
\begin{equation} 
\nu(\{\top^{(0)}\})=\nu(\{\top^{(1)}\})=\emptyset \subset \{1\}=\nu\left(\{\top^{(0)}\}\vee \{\top^{(1)}\}\right)
\end{equation} 
so the composition of the two lotteries is not explainable. 
In Figures \ref{subfig: Ellsberg non-explainable, lower poset}-\ref{subfig: Ellsberg non-explainable, valuation}, we provide a graphical representation of the previous argument based on Hasse diagrams \citep[Sect. 1.15]{Davey2002}, denoting $\ell_{0}:=\{\top^{(0)}\}$ and $\ell_{1}:=\{\top^{(1)}\}$ to stress the link to the two lotteries in Ellsberg's model.

\begin{figure}
\subfigure[Using lower meta-states]
{
\begin{tikzpicture}
\node (v00) at (-2,-4.5) {$\left(\emptyset\right)^{(0)}$};
\node (v10) at (2,-4.5) {$\left(\emptyset\right)^{(1)}$};

\node (v01) at (-2,-3) {$\left(\ell_{0}\right)$};
\node (v0010) at (0,-3) {$\left(\emptyset,\emptyset\right)$};
\node (v11) at (2,-3) {$\left(\ell_{1}\right)$};

\node (v0110) at (-2,-1.5) {$\left(\ell_{0},\emptyset\right)$};
\node (v0011) at (2,-1.5) {$\left(\emptyset,\ell_{1}\right)$};

\node (v0111) at (0,0) {$\left(\ell_{0},\ell_{1}\right)$};

\draw  (v00) edge (v01);
\draw  (v10) edge (v11);
\draw  (v0010) edge (v0110);
\draw  (v0010) edge (v0011);
\draw  (v0110) edge (v0111);
\draw  (v0011) edge (v0111);
\draw  (v00) edge [dashed] (v0010);
\draw  (v10) edge [dashed] (v0010);
\draw  (v01) edge [dashed] (v0110);
\draw  (v11) edge [dashed] (v0011);
\label{subfig: Ellsberg non-explainable, lower poset}
\end{tikzpicture}
}
\hfill
\subfigure[Using $\nu$-meta-states]
{
\begin{tikzpicture}
\node (v00) at (-2,-4.5) {$\left(\emptyset\right)^{(0)}$};
\node (v10) at (2,-4.5) {$\left(\emptyset\right)^{(1)}$};

\node (v01) at (-2,-3) {$\left(\ell_{0}\right)$};
\node (v0010) at (0,-3) {$\left(\emptyset,\emptyset\right)$};
\node (v11) at (2,-3) {$\left(\ell_{1}\right)$};

\node (v0110) at (-2,-1.5) {$\left(\ell_{0},\emptyset\right)$};
\node (v0011) at (2,-1.5) {$\left(\emptyset,\ell_{1}\right)$};

\node (v0111) at (0,0) {$\left(\ell_{0},\ell_{1}\right)$};

\node(vb) at (4,-4.5) {$\emptyset$};
\node(vt) at (4,0) {\{1\}};

\draw  (v00) edge (v01);
\draw  (v10) edge (v11);
\draw  (v0010) edge (v0110);
\draw  (v0010) edge (v0011);
\draw  (v0110) edge (v0111);
\draw  (v0011) edge (v0111);
\draw  (v00) edge (v0010);
\draw  (v10) edge (v0010);
\draw  (v01) edge (v0110);
\draw  (v11) edge (v0011);
\draw  (vb) edge (vt);

\begin{scope}[dotted]
\draw[->]  (v00) -- (vb);
\draw[->]  (v10) -- (vb);
\draw[->]  (v01) -- (vb);
\draw[->]  (v11) -- (vb);

\draw[->]  (v0010) -- (vt);
\draw[->]  (v0110) -- (vt);
\draw[->]  (v0011) -- (vt);
\draw[->]  (v0111) -- (vt);
\end{scope}
\label{subfig: Ellsberg non-explainable, valuation}
\end{tikzpicture}
}
\label{fig: Ellsberg non-explainable}
\caption{Representations of ambiguity in Ellsberg's three-colour model using the two types of meta-states. Dashed lines (left figure) distinguish pairs in the poset $\varrho^{\star}$ that do not belong to the poset $\varrho_{\star}$. Dotted lines (right figure) depict the map $\nu$ linking $\varrho^{\star}$ to the Boolean algebra $V_{\text{exp}}:=\mathtt{B}^{(1)}$.}
\end{figure}
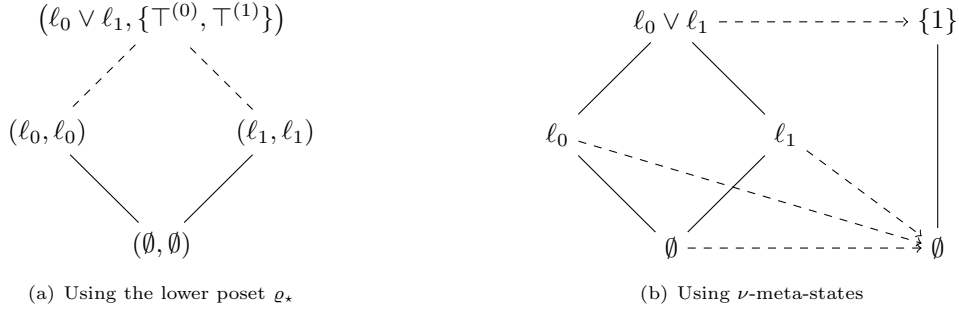

We point out that a different view on an analogous phenomenon was given in \cite[Sect. 7]{Angelelli2017}, where different orders for quantities characterising physical subsystems emerge as a consequence of different choices of normalisations. In fact, (\ref{eq: violation Sure-Thing principle}) follows from a different attribution of the ``ground energy'', or minimal value, here interpreted as the intersection of the alternatives in each scenario, i.e., $\emptyset$ for the set of alternatives $\{\pi_{0,a},\pi_{0,b}\}$ and $\{\texttt{Y}\}$ for the set $\{\pi_{1,a},\pi_{1,b}\}$.
These two different choices represent unrelated normalisations, which open the way to incompatibility of preferences (opposite orders) between the two scenarios. We can better specify this observation by linking the decision contexts that generate the poset in Figure \ref{subfig: Ellsberg non-explainable, lower poset}, i.e., the maximal elements of $\varrho_{\star}$, to Boolean algebras. Specifically, we map the Boolean algebra $\mathtt{B}^{(2)}$ to lotteries $\ell_{u}$, for both $u\in\{0,1\}$, and the Boolean algebra $\mathtt{B}^{(3)}$ to their combination $\ell_{0}\vee\ell_{1}$: 
\begin{eqnarray}
\ell_{0} & \mapsto & \mathcal{B}_{1}:=\left(\wp\{\mathtt{R},\mathtt{B}\},\cup,\cap,\cdot \rightarrow \emptyset,\{\mathtt{R},\mathtt{B}\},\emptyset\right), 
\nonumber \\ 
\ell_{1} & \mapsto & \mathcal{B}_{2}:=\left(\left\{\{\mathtt{Y}\},\{\mathtt{R},\mathtt{Y}\},\{\mathtt{B},\mathtt{Y}\},\{\mathtt{R},\mathtt{Y},\mathtt{B}\}\right\},\cup,\cap,\cdot \rightarrow\{\mathtt{Y}\},\{\mathtt{R},\mathtt{Y},\mathtt{B}\},\{\mathtt{Y}\}\right),
\nonumber \\ 
\ell_{0}\vee\ell_{1} & \mapsto & \mathcal{B}_{1,2}:=\left(\wp\{\mathtt{R},\mathtt{Y},\mathtt{B}\},\cup,\cap,\cdot \rightarrow \emptyset,\{\mathtt{R},\mathtt{Y},\mathtt{B}\},\emptyset\right)
\label{eq: extending Boolean algebras}
\end{eqnarray}
where we have used the expression $\cdot\rightarrow \emptyset$ for the complement $\cdot^{\mathtt{C}}$. The ambiguity in Ellsberg's model entails the lack of a combination of the two Boolean algebras $\mathcal{B}_{1}$ and $\mathcal{B}_{2}$ to get $\mathcal{B}_{1,2}$. This combination would be feasible if we could distinguish the two algebras and associate them with sub-structures of $\mathcal{B}_{1,2}$. However, meta-reasoning (lower poset $\varrho_{\star}$ in Figure \ref{subfig: Ellsberg non-explainable, lower poset}, $\nu$-meta-states in Figure \ref{subfig: Ellsberg non-explainable, valuation}) does not allow for such a distinction. In this way, we get another instance of multiple implications (here, $\cdot\rightarrow \emptyset$ and $\cdot\rightarrow \{\mathtt{Y}\}$) already considered in Example \ref{exa: lack of knowledge for complementation}, as well as the non-trivial effect of a ``ground energy'' labelling subsystems (here, $\emptyset$ and $\{\mathtt{Y}\}$ in the aforementioned implications) as in \citep{Angelelli2017}.

\subsection{Wigner's model and data observability}
\label{subsec: Wigner effect and data observability}

While the labelling of contexts refers to lotteries in Ellsberg's model, in the Wigner's Friend scenario, it represents the two potential changes in the friend's state implied by the measurement, which are unknown to Wigner. As in the case of data-driven strategies, an agent is able to observe data (outcome of measurement in the first case, - big - data in the second one) that a super-observer (Wigner in the first case, a human agent in the second one) cannot. 

Even for the Wigner's Friend scenario, before defining the formal correspondence with our framework (Subsection \ref{subsubsec: representing uncertainty on data observability}), we briefly discuss the source of uncertainty about features extracted from non-explainable approaches using artificial agents.

\subsubsection{Preliminary discussion}
\label{subsubsec: Wigner preliminary discussion}

Let us consider an order relation on $\mathtt{f}_{(\mathrm{A})}:=\{\mathtt{f},\mathtt{f}_{0},\mathtt{f}_{1}\}$ given by the inclusion of the indices. Specifically, we have $\mathtt{f}\preceq_{\mathrm{data}}\mathtt{f}_{0}$ and $\mathtt{f}\preceq_{\mathrm{data}}\mathtt{f}_{1}$. The label $\mathtt{f}$ refers to the word ``factor'' (or feature), namely, a relevant attribute defining the decision context based on the observed data. The condition $\mathtt{f}\preceq_{\mathrm{data}}\mathtt{f}_{0}$ means that $\mathtt{f}$ does not identify a decision context, while $\mathtt{f}_{0}$ does and, hence, is less ambiguous. Note the analogy with the urn model described in the previous subsection: $\mathtt{f}_{0}$ and $\mathtt{f}_{1}$ could represent two distinct decision contexts $\ell_{0}$ and $\ell_{1}$, resulting in two opposite orders.

The set $\mathtt{f}_{(\mathrm{A})}$ refers to the direct observation of data carried out by the artificial agent. So we move to a second representation $\mathtt{F}_{(\mathrm{H})}:=\left\{ \mathtt{F}_{\{0\}},\mathtt{F}_{\{1\}},\mathtt{F}_{\{\emptyset\}},\mathtt{F}_{\emptyset}\right\} $ to assess the knowledge possessed by the human agent about the AI's decisions. Specifically, the element $\mathtt{F}_{\{\emptyset\}}$ recognises that the trained AI algorithm is in a defined but unknown decision context, and we consider the relations  $\mathtt{F}_{\emptyset}\preceq_{\mathrm{AI}}\mathtt{F}_{\{\emptyset\}}\preceq_{\mathrm{AI}}\mathtt{F}_{\{u\}}$ for both $u\in\{0,1\}$, where $\mathtt{F}_{\{\emptyset\}}$ is interpreted as ``\textit{the human agent knows that the AI knows the decision context}'', while $\mathtt{F}_{\emptyset}$ is interpreted as ``\textit{the human agent knows that the AI does not know the decision context}''. The subscript $\preceq_{\mathrm{AI}}$ clarifies that the comparison refers to the AI's knowledge state. 

As a consequence of the training with data, the AI updates its initial state to align with them. This update leads to the definition of a meta-state for the human agent, which reflects the changes in the AI's knowledge. Specifically, the knowledge of the AI's training prompts the human agent's knowledge to change from $\mathtt{F}_{\emptyset}$ to $\mathtt{F}_{\{\emptyset\}}$. This update acknowledges the alignment of the AI's outcomes with a feature in the (big) data, but the human agent remains unaware of the specific latent feature. 

The relational structure defined by $\mathtt{F}_{\mathrm{(H)}}$ can be encoded using the function $\iota$ introduced in (\ref{eq: micro-macro transformation}) to provide an instance of inequivalent knowledge representations. Specifically, we observe that 
\begin{equation}
\iota\left(\emptyset\right)=\emptyset,\quad\iota(\{0\})=\iota(\{1\})=\{\emptyset\},\quad\iota(\{0,1\})=\{\emptyset,\{0\},\{1\}\}.
\label{eq: Wigner micro-macro}
\end{equation}
Then, the update from $\mathtt{f}$ to $\mathtt{f}_{u}$ for some $u\in\{0,1\}$ prompts the update from $\mathtt{F}_{\iota(\emptyset)}$ to $\mathtt{F}_{\iota(\{u\})}$. We can describe the human agent's inability to explain the AI's outcome through the set difference $\Delta:=\iota(\{0,1\})\setminus \iota(\{u\})$ as a means to represent the divergence between the full access to knowledge about the AI's decision contexts $\{0\}\vee \{1\}=\{0,1\}$ (in $\left(\wp(\{0,1\},\subseteq\right)$) and the actual knowledge $\{u\}$. In this interpretation, from $\Delta\neq \emptyset$, we can say that the states $\mathtt{F}_{U}$ with $U\in\Delta$ are not accessible to the human agent. 

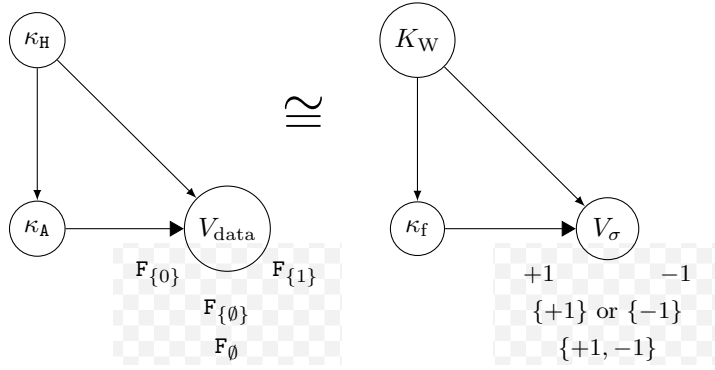
\begin{figure} 
\begin{center}
\begin{tikzpicture}

\path [pattern=checkerboard,pattern color=black!5] (-1,-2.8) rectangle (2,-1.2);
\path [pattern=checkerboard,pattern color=black!5] (4,-2.8) rectangle (7,-1.2);

\node (v1) at (-2,1.5) [style={draw,shape=circle}] {$\kappa_{\mathtt{H}}$};
\node (v2) at (-2,-1) [style={draw,shape=circle}] {$\kappa_{\mathtt{A}}$};
\node (v3) at (0.5,-1) [style={draw,shape=circle}] {$V_{\mathrm{data}}$};
\draw [-latex] (v1) edge (v2);
\draw [-latex] (v1) edge (v3);
triangle 60
\draw [-triangle 60] (v2) edge (v3);
\node (v7) at (0.5,-2.6)   {\scriptsize{$\texttt{F}_{\emptyset}$}};
\node (v8) at (0.5,-2.1) {\scriptsize{$\texttt{F}_{\{\emptyset\}}$}};
\node (v9) at (-0.4,-1.6) {\scriptsize{$\texttt{F}_{\{0\}}$}};
\node (v10) at (1.4,-1.6) {\scriptsize{$\texttt{F}_{\{1\}}$}};

\node (v4) at (3,1.5) [style={draw,shape=circle}] {$K_{\mathrm{W}}$};
\node (v5) at (3,-1) [style={draw,shape=circle}] {$\kappa_{\mathrm{f}}$};
\node (v6) at (5.5,-1) [style={draw,shape=circle}] {$V_{\mathrm{\sigma}} $};
\node (v11) at (5.5,-2.6) {\scriptsize{$\{+1,-1\}$}};
\node (v12) at (5.5,-2.1) {\scriptsize{$\{+1\}\text{ or }\{-1\}$}};
\node (v13) at (4.6,-1.6) {\scriptsize{$+1$}};
\node (v14) at (6.4,-1.6) {\scriptsize{$-1$}};
\draw [-latex] (v4) edge (v5);
\draw [-latex] (v4) edge (v6);
triangle 60
\draw [-triangle 60] (v5) edge (v6);
\node at (1.5,0.5) {\huge{$\cong$}};
\end{tikzpicture}
\end{center}
\caption{Diagrammatic representation of a Wigner's friend scenario describing a high-level interaction of human and artificial agents in a data-driven scenario.}
\label{fig: Wigner's friend} 
\end{figure}

This argument, which is graphically depicted in Figure \ref{fig: Wigner's friend}, is a basis for the specification of the formalism designed in this work for the Wigner's Friend scenario, as presented in the following subsection. 

\vspace{0.5cm}

\subsubsection{Representing uncertainty about data observability}
\label{subsubsec: representing uncertainty on data observability}

For the dimension $V_{\mathrm{data}}$, we choose a base set $\wp(\{\top\})$ to align with the original Wigner's Friend scenario. This encoding captures the effect of the spin measurement by Wigner's Friend as a transition from the two-dimensional space with basis $\{s_{+1},s_{-1}\}$ of $\mathbb{C}^{2}$ to one of the one-dimensional spaces (with basis $\{s_{+1}\}$ or $\{s_{-1}\}$, respectively). Therefore, we define $V_{\mathrm{data}}:=\left(\wp(\{\top\}),\subseteq\right)$ to represent the knowledge (observation or measurement) of a relevant feature (polarisation) that allows value extraction from data (measured spin). Other definitions of $V_{\mathrm{data}}$ can be considered too, but it is worth noting that this choice also connects to the representation of Ellsberg's model in the previous subsection through the association of $\top$ with the knowledge of the ground energy, which distinguishes the inferences made in the two lotteries. 

Data are observable for the AI (Wigner's Friend) but not for the human agent (Wigner); only the knowledge of the existence of an outcome observed by the AI (e.g., the conclusion of the training phase) is available to the human agent. Assuming that no other knowledge source besides data is needed to define the state of the AI, we set $\varrho_{\mathrm{A}}:=V_{\mathrm{data}}$. While the acknowledgement of data value is given by $\pi_{\mathrm{data}}(\kappa_{\mathrm{H}})$, the acknowledgement of the value of the AI in $\kappa_{\mathrm{H}}$ is represented by the component $\pi_{\mathrm{AI}}(\kappa_{\mathrm{H}})$ along the dimension $V_{\mathrm{AI}}:=\left(\wp(\varrho_{\mathrm{A}}),\subseteq\right)$. This abstracts the queries (measurements) $\wp(\kappa_{\mathrm{A}})$ that the human agent (Wigner) can ask the AI (Wigner's Friend) with state $\kappa_{\mathrm{A}}$. 

As a consequence of the actual observation of the outcome in the data-driven scenario, the states $\kappa_{\mathrm{A}}$ and $\kappa_{\mathrm{H}}$ are updated to encompass the existence of $\top$. The AI gains complete information about the relevant data dimension, leading to a change from $\pi_{\mathrm{data}}(\kappa_{\mathrm{A}})=\emptyset$ to a new state $\kappa'_{\mathrm{A}}$ with $\pi_{\mathrm{data}}(\kappa'_{\mathrm{A}})=\{\top\}$. We can express this update in accordance with Definition \ref{def: explainability} by introducing $\psi_{\mathrm{A}}:=\left(\{\top\}_{\mathrm{data}}\right)$ and using the composition   
\begin{equation}
\kappa_{\mathrm{A}}=\left(\emptyset_{\mathrm{data}}\right)\mapsto\kappa'_{\mathrm{A}}:=\kappa_{\mathrm{A}}\vee\psi_{\mathrm{A}}=\left(\{\top\}_{\mathrm{data}}\right).
\label{eq: Wigner, update A}
\end{equation}
On the other hand, the realisation of the training of the AI algorithm prompts a change in the knowledge state of the human agent; consistently with the transition $\kappa_{\mathrm{A}}\mapsto\kappa'_{\mathrm{A}}$, we describe 
\begin{equation}
\kappa_{\mathrm{H}}=\left(\emptyset_{\mathrm{data}},\wp(\emptyset)_{\mathrm{AI}}\right)\mapsto\kappa'_{\mathrm{H}}:=\left(\emptyset_{\mathrm{data}},\wp(\{\top\})_{\mathrm{AI}}\right)
\label{eq: Wigner, update H}
\end{equation}
which means that the human agent knows that the AI is aligned with the data provided for the training but is unable to directly query them. To formulate this limitation, we can express the transition from $\kappa_{\mathrm{H}}$ to $\kappa'_{\mathrm{H}}$ by introducing $\psi_{\mathrm{H}}:=\left(\emptyset_{\mathrm{data}},\{\{\top\}\}_{\mathrm{AI}}\right)$ and looking at the explainability of the update $\kappa_{\mathrm{H}}\vee \psi_{\mathrm{H}}$ by $\nu$-meta-states. As discussed in Subsection \ref{subsubsec: Wigner preliminary discussion}, we can use $\iota$ to generate $\nu$-meta-states; an equivalent choice to analyse this scenario, where the update from $\kappa_{\mathrm{H}}$ to $\kappa'_{\mathrm{H}}$ does not affect the data dimension, is $\nu(\kappa_{\mathrm{H}}):=\iota\circ \pi_{\mathrm{AI}}(\kappa_{\mathrm{H}})$ where $\mathrm{AI}\in\underline{\kappa_{\mathrm{H}}}$, and $\emptyset$ otherwise. We find
\begin{eqnarray}
(\kappa_{\mathrm{H}},\iota(\kappa_{\mathrm{H}}))\vee(\psi_{\mathrm{H}},\iota(\psi_{\mathrm{H}})) & = & \left(\kappa_{\mathrm{H}}\vee\psi_{\mathrm{H}},\iota(\kappa_{\mathrm{H}})\vee\iota(\psi_{\mathrm{H}})\right)\nonumber \\
& = & \left(\left(\emptyset_{\mathrm{data}},\wp(\{\top\})_{\mathrm{AI}}\right),\{\emptyset\}\right)\nonumber \\
& \neq & \left(\left(\emptyset_{\mathrm{data}},\wp(\{\top\})_{\mathrm{AI}}\right),\left\{ \emptyset,\{\emptyset\},\{\{\top\}\}\right\} \right) \nonumber \\ 
& = &(\kappa_{\mathrm{H}}\vee\psi_{\mathrm{H}},\iota(\kappa_{\mathrm{H}}\vee\psi_{\mathrm{H}})).
\label{eq: Wigner, non-explainable update}
\end{eqnarray}
So the update leading to $\kappa'_{\mathrm{H}}$ is not explainable based on the previous definitions.

\section{Discussion on implications for the assessment of data-driven strategies}
\label{sec: scope of the work}

As mentioned in the Introduction and Section \ref{sec: value uncertainty scenarios}, business performance, as an indicator of the value associated with a big data-driven strategy, can be seen as a consequence of actions implemented in line with the strategic value generated through big data analysis. This stage, which can be defined as \textit{big data exploitation}, is itself a consequence of a pre-process aimed at interpreting and capitalising information extracted from data, which is referred to as \textit{big data capitalization} \citep{YlijokiPorras2019, Wu2022707}. Based on findings in the literature, the dimensional definition of value is limited to the final phase of big data exploitation, as its principal purpose is to specialise the different types of value associated with big data in the measurement of the generated value. This measurement is realised by comparing the observed and estimated value indicators, which provides a criterion to determine the success or failure of the implemented data-driven strategy.

The proposed framework acts as an additional layer that embeds value dimensions within a formal knowledge structure and, hence, encompasses specific types of uncertainty (Subsection \ref{subsec: sources of uncertainty in data-driven strategies}) within the capitalization phase. In practice, these types of uncertainty undermine the data-driven strategy from the beginning due to the lack of proper assessments of useful information when data are in their raw state. This aspect is critical since the definition of data-driven strategies often takes place before raw data acquisition, when \textit{a priori} evaluations of information or strategic value are not available \citep{Manyika}. On the contrary, the structures to represent knowledge, in particular inner states, could be implemented within the whole value transformation process to highlight potential inconsistencies, redundancies, or a lack of representative indicators. In this way, the present model could be adapted for strategy assessment over the different stages of data capitalization and exploitation. In addition, it could enhance the applicability of in-use big data frameworks by promoting their proper adoption and integration within specific phases of the value-creation processes. 

The analysis conducted by \citet{Ashton2007} points out the need to understand and consider casual linkages in these indicators, as they may create redundancies and distortions in aggregate measures such as the \textit{value creation index} (VCI). The VCI introduces new categories of information (e.g., innovation, alliances, and technology), which are combined and weighted along with firm performance \citep{Ashton2007}. In this regard, the integration of inner state representations within the Big Data Value Chain is a means to formalise and, if needed, update the knowledge about the processes that lead to value creation. Specifically, the cyclical assessment of potential inconsistencies that require updating a knowledge representation may return evidence about the resource usage and skills needed at the different assessment stages, which is a premise for the study of (causal) relations among them. 

The focus on the explainability of inner states' updates also matches the need to adapt the assessment methodologies from the capitalization to the exploitation phases. Indeed, value measurement in terms of business performance can benefit from both classical forecasting models and advanced analytic methods \citep{Negro2022}, which allow for the estimation of probability laws to obtain informative statistics and indices. However, this focus may overlook the quantification of non-financial and organisational indicators. This requires the introduction of complementary methods to assess these value attributes. Our proposal fits into the design of such methods that, starting with knowledge representations, assess their compatibility and the need to update the dimensional structure. In line with current research streams that explore the measurement of non-classical forms of uncertainty in socio-economic and psychometric settings (Subsection \ref{subsec: evidence of non-classical behaviour in cognition and self-assessment}), such methods formalise a type of configural invariance between different frameworks and studies (in the sense discussed in Subsection \ref{subsec: dimensional frameworks as state transitions}). 

The conditions expressed in Section \ref{sec: model proposal} are grounded in the theory of extended orders, which go beyond the classical Boolean structures underlying probabilistic models. In this way, we strengthen a common foundational basis for fuzzy logic \citep{DellaStella2012} and for the study of inequivalent representations of statistical systems \citep{Angelelli2017}, as discussed in Subsection \ref{sec: properties of knowledge frames and their relations to data-driven strategies}.
In fact, the attention paid to element- and set-based representations is part of a broader investigation that explores reduction to or deviation from classical set membership through geometric models. Specifically, a geometric realisation of the operational structure introduced in Subsection \ref{subsec: Comparison and composition} arises from a limiting procedure for set functions, where a set-element correspondence derives from combinatorial \citep[Sect. 6-7]{Angelelli2018} or algebraic constraints \citep{Angelelli2019}. Other types of uncertainty can be explored from a geometric standpoint, including imprecise probabilities \citep[Sect. 2.2]{Cuzzolin2020} and factor indeterminacy in multivariate methods \citep[pp. 430-431]{Rigdon2019}. The latter shares a feature with the type of metrological uncertainty discussed in this work, namely, the multiplicity of models compatible with the same accessible or observed information. Indeed, although different in nature, both types of indeterminacy rely on families of model transformations that preserve a given structure. In \citep{Rigdon2019}, they generate different solutions consistent with the same covariance structure, while in our setting, we deal with order-preserving mappings with different operational structures (\ref{eq: tropical addition to poset}).

In conclusion, this formalism lends itself to the generation of qualitative and quantitative criteria for meta-reasoning about knowledge, which may support reliable maturity assessments (Subsection \ref{subsec: evidence of non-classical behaviour in cognition and self-assessment}). Further study should be devoted to the specification of our approach for the design of adaptive maturity models in line with the dynamics of capabilities and technological adoption.

\section{Conclusion and future work}
\label{sec: conclusion}

This work has laid the basis for a deeper investigation of knowledge uncertainty in data-driven strategies, which are becoming a dominant component of technological innovation with significant effects on socio-economic systems \citep{Ndou2019}. The contribution started by identifying different manifestations of uncertainty that may affect data-driven strategies, which have to be included in the intermediate and final evaluations of innovation initiatives for their proper analysis. 

Here, we focused on the explainability of knowledge updates; future work will explore the structures to express and investigate the explainability of knowledge \emph{states}. The relation between structural (logical and algebraic) and information-theoretic notions should be explored in more depth to exploit both qualitative and quantitative approaches for assessing uncertainty in epistemic representations. Specifically, normalisation's role in representing information-theoretic inequivalence (Subsection \ref{subsec: uncertainty on information measures}) and the Ellsberg model (Subsection \ref{subsubsec: ambiguity and non-explainable updates}) can be translated into an information-geometric setting \citep[Sect. 5]{Angelelli2021}, prompting dedicated analyses within other entropy-based statistical models \citep{Carpita2017} and fuzzy techniques \citep{Ciavolino2014,Ciavolino2016}.

In this way, we envisage practical advantages in the design of measurement tools to assess business maturity in the context of big data. Maturity has naturally been linked to business value through the assumption that an organisation with a high level of maturity has a greater chance of turning potential value into created value. At the same time, maturity models and dimensional value models share many of the aspects that have been explored in this discussion. Future applications will define assessment questionnaires suited to the analysis of interactions between human agents and technologies for data-driven initiatives. Along with a chosen methodological architecture in terms of dimensions, these assessment tools should envisage the occurrence of multiple representations of the same latent construct with incompatible behaviours (e.g., different qualitative features of relations within the structural model). 

A final aspect to consider, beyond the scope of this paper, is the critical analysis of the non-monotonic relation between data features and the value that can be generated. Having more data (higher volume) is not always synonymous with getting a higher value. From a perspective in which data are resources, we should look at the extent to which data, information, and knowledge representations could faithfully be represented as resources, or, instead, they require a multi-actor view. In this direction, \cite{Gervasi_2023_(1)} discussed how big data value chain models can be combined with new data governance models, such as the Data Mesh \cite{Dehghani_Only_2020}.
The presented framework is likely to fit into this multi-actor scheme, where AI is an agent and is part of the tools that can be used by a Technology Mesh to combine data from different domains \citep{Gervasi_2023_(1)}. In this way, data-driven strategies should be considered on a different ground with respect to classical paradigms in software engineering. The variety of effects that could be generated by data-driven strategies and the use of AI tools, as we discussed, should be incorporated within management processes, as they need to support organisations in increasing their awareness of data-driven strategies and the reliability of generated value measurements.

\bibliography{sn-bibliography}

\end{document}